\documentclass[twoside]{article}

\usepackage[accepted]{aistats2019}

\usepackage[utf8]{inputenc} % allow utf-8 input
\usepackage[T1]{fontenc}    % use 8-bit T1 fonts
\usepackage{hyperref}       % hyperlinks
\usepackage{url}            % simple URL typesetting
\usepackage{booktabs}       % professional-quality tables
\usepackage{amsfonts}       % blackboard math symbols
\usepackage{nicefrac}       % compact symbols for 1/2, etc.
\usepackage{microtype}      % microtypography
\usepackage[pdftex]{graphicx}
\usepackage{flushend}

\newcommand{\argmax}{\mathop{\rm arg~max}\limits}

\begin{document}

\twocolumn[

\aistatstitle{Adaptive Ensemble Prediction for Deep Neural Networks based on Confidence Level}

\aistatsauthor{ Hiroshi Inoue <inouehrs@jp.ibm.com> }

\aistatsaddress{ IBM Research - Tokyo } ]

\begin{abstract}
Ensembling multiple predictions is a widely used technique for improving the accuracy of various machine learning tasks. One obvious drawback of ensembling is its higher execution cost during inference. In this paper, we first describe our insights on the relationship between the probability of prediction and the effect of ensembling with current deep neural networks; ensembling does not help mispredictions for inputs predicted with a high probability even when there is a non-negligible number of mispredicted inputs. This finding motivated us to develop a way to adaptively control the ensembling. If the prediction for an input reaches a high enough probability, i.e., the output from the softmax function, on the basis of the confidence level, we stop ensembling for this input to avoid wasting computation power. We evaluated the adaptive ensembling by using various datasets and showed that it reduces the computation cost significantly while achieving accuracy similar to that of static ensembling using a pre-defined number of local predictions. We also show that our statistically rigorous confidence-level-based early-exit condition reduces the burden of task-dependent threshold tuning better compared with naive early exit based on a pre-defined threshold in addition to yielding a better accuracy with the same cost.
\end{abstract}

\section{Introduction}

The huge computation power of today's computing systems, equipped with GPUs, special ASICs, FPGAs, or multi-core CPUs, makes it possible to train deep networks by using tremendously large datasets. Although such high-performance systems can be used for training, actual inference in the real world may be executed on small devices, such as a handheld devices or an embedded controller. %, which have much smaller computation power and energy supply than the large systems used for training the network. 
Hence, various techniques (such as \cite{Hinton15} and \cite{Han16}) for achieving a high prediction accuracy without increasing computation time have been studied to enable more applications to be deployed in the real world.
% To reduce the computation costs in the inference phase, \cite{Hinton15} created a smaller network for deployment by distilling the knowledge from an ensemble of multiple models. \cite{Han16} also targeted deployment for small (mobile) devices and showed that large networks can be significantly compressed after training by pruning unimportant connections and by quantizing each connection. 
% Recently, graph compilers and optimizers, such as NNVM/TVM (\cite{Chen18}) and Glow (\cite{Rotem18}), were developed to make small and efficient executables suitable for handheld devices from trained networks.

Ensembling multiple predictions is a widely used technique for improving the accuracy of various machine learning tasks (e.g., \cite{Hansen90}, \cite{Zhou02}) at the cost of more computation power. In image classification tasks, for example, accuracy is significantly improved by ensembling local predictions for multiple patches extracted from an input image to make a final prediction. Moreover, accuracy is further improved by using multiple networks trained independently to make local predictions. \cite{Krizhevsky12} averaged 10 local predictions using 10 patches extracted from the center and the 4 corners of input images with and without horizontal flipping in their Alexnet paper. They also used up to seven networks and averaged the prediction to increase the accuracy. GoogLeNet by \cite{Szegedy15} averaged up to 1,008 local predictions by using 144 patches and 7 networks. % In their paper, they reported that averaging 1,008 predictions reduced the top-5 error rate of ImageNet classification task by 3.45\% whereas averaging 10 predictions with one model reduced the error by 0.92\% compared with the baseline prediction without ensembling. 
In some ensemble methods, meta-learning during training to learn how to best mix multiple local predictions from the networks is used (e.g., \cite{Tekin16}). In Alexnet or GoogLeNet papers, however, significant improvements have been obtained by just averaging local predictions without meta-learning. In this paper, we do not use meta-learning either.

Although the benefits of ensemble prediction are quite significant, one obvious drawback is its higher execution cost during inference. If we make a final prediction by ensembling 100 predictions, we need to make 100 local predictions, and, hence, the execution cost will be 100 times as high as that without ensembling. This higher execution cost limits the real-world use of ensembling, especially on small devices, even though using it is almost the norm to win image classification competitions that emphasize prediction accuracy. 

\begin{figure*}[t]
\begin{center}
  \begin{tabular}{c}
    \begin{minipage}{0.5\hsize}  
      \includegraphics[width=10.5cm,trim=1cm 6cm 0 0]{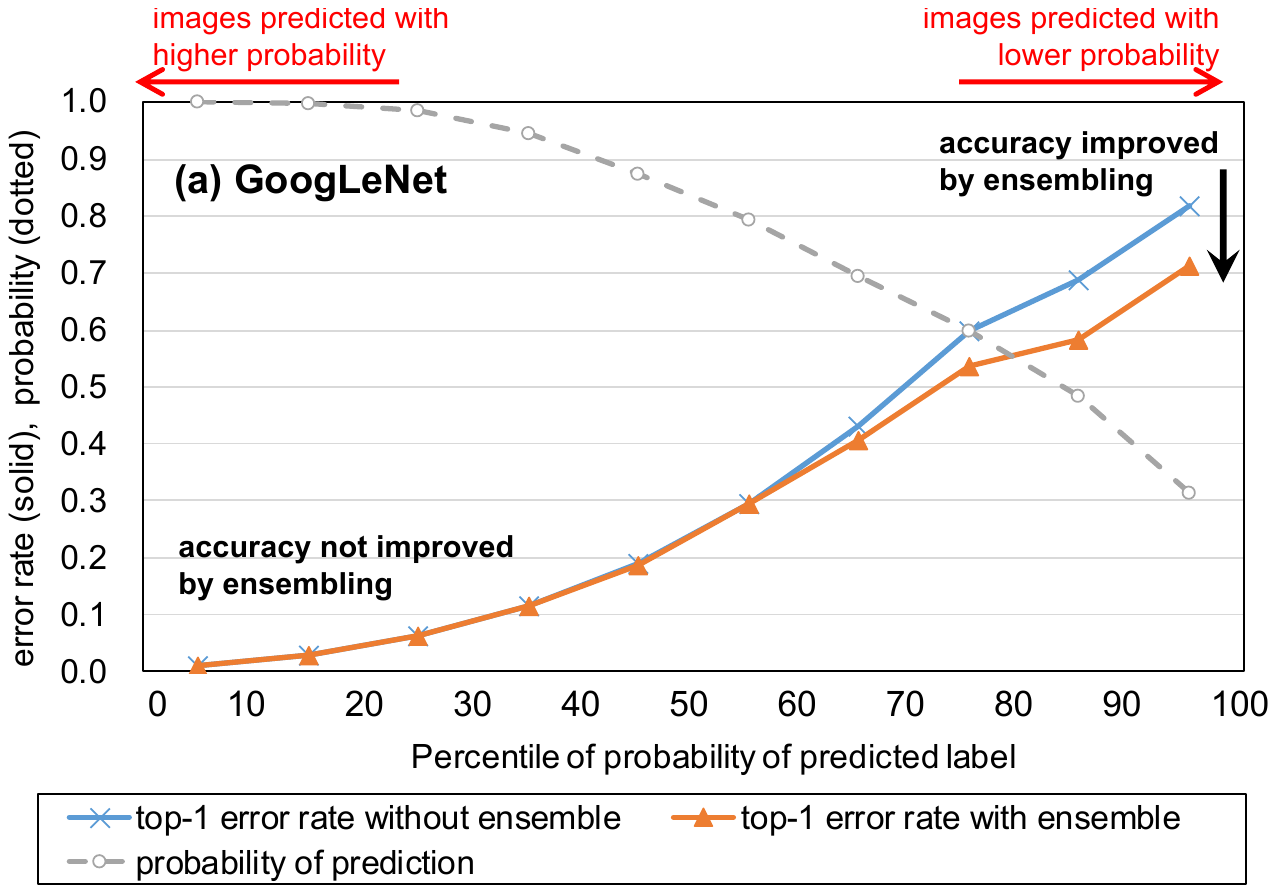}
    \end{minipage}
    \begin{minipage}{0.5\hsize}  
      \includegraphics[width=10.5cm,trim=1cm 6cm 0 0]{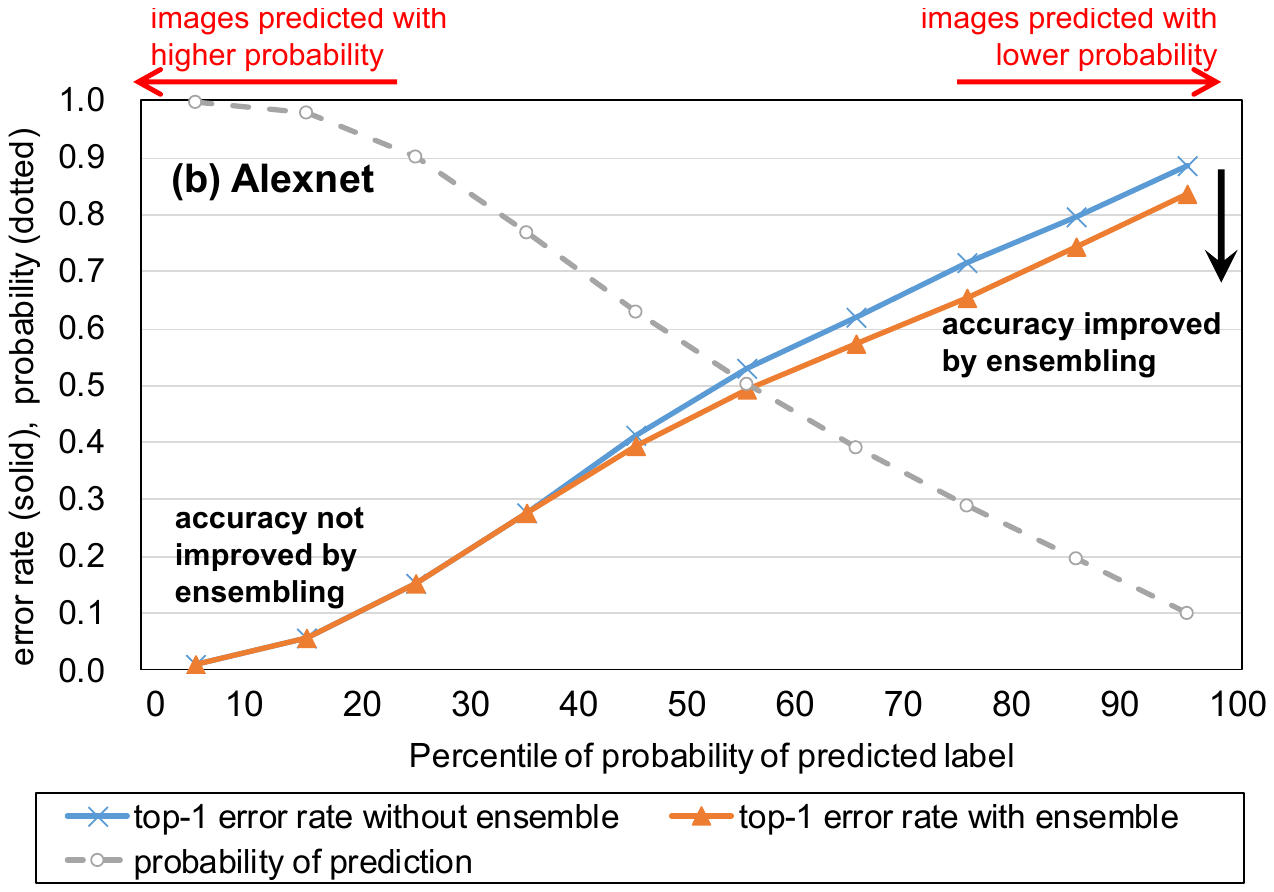}
    \end{minipage}
  \end{tabular}
  \caption{Improvements made by ensembling and probabilities of predictions for ILSVRC 2012 validation set. X-axis shows percentile of probability of first local predictions from high (left) to low (right). Ensembling reduces error rates for inputs with low probabilities but does not affect inputs with high probabilities.}

  \begin{tabular}{c}
    \begin{minipage}{0.5\hsize}  
      \includegraphics[width=10cm,trim=1.8cm 7cm 0 0]{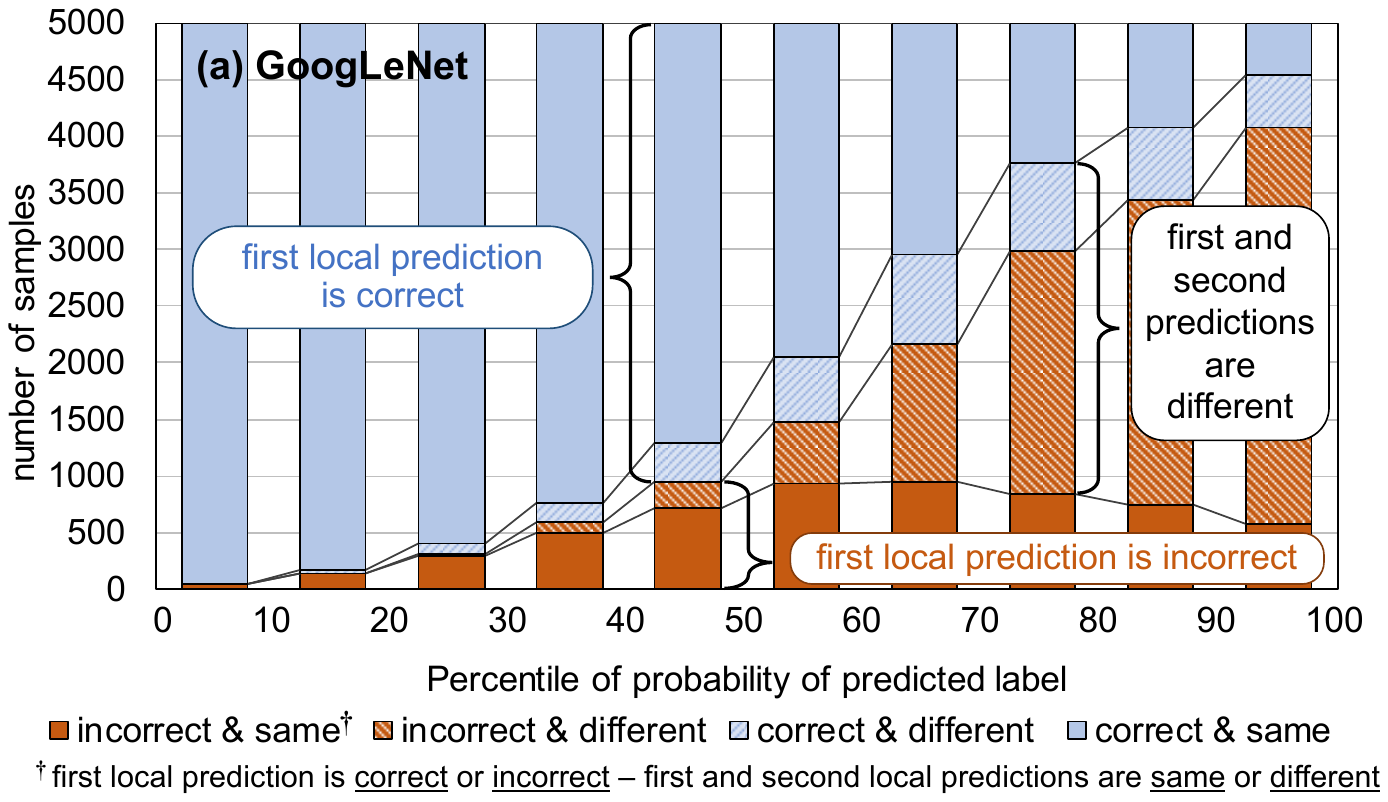}
    \end{minipage}
    \begin{minipage}{0.5\hsize}  
      \includegraphics[width=10cm,trim=1.8cm 7cm 0 0]{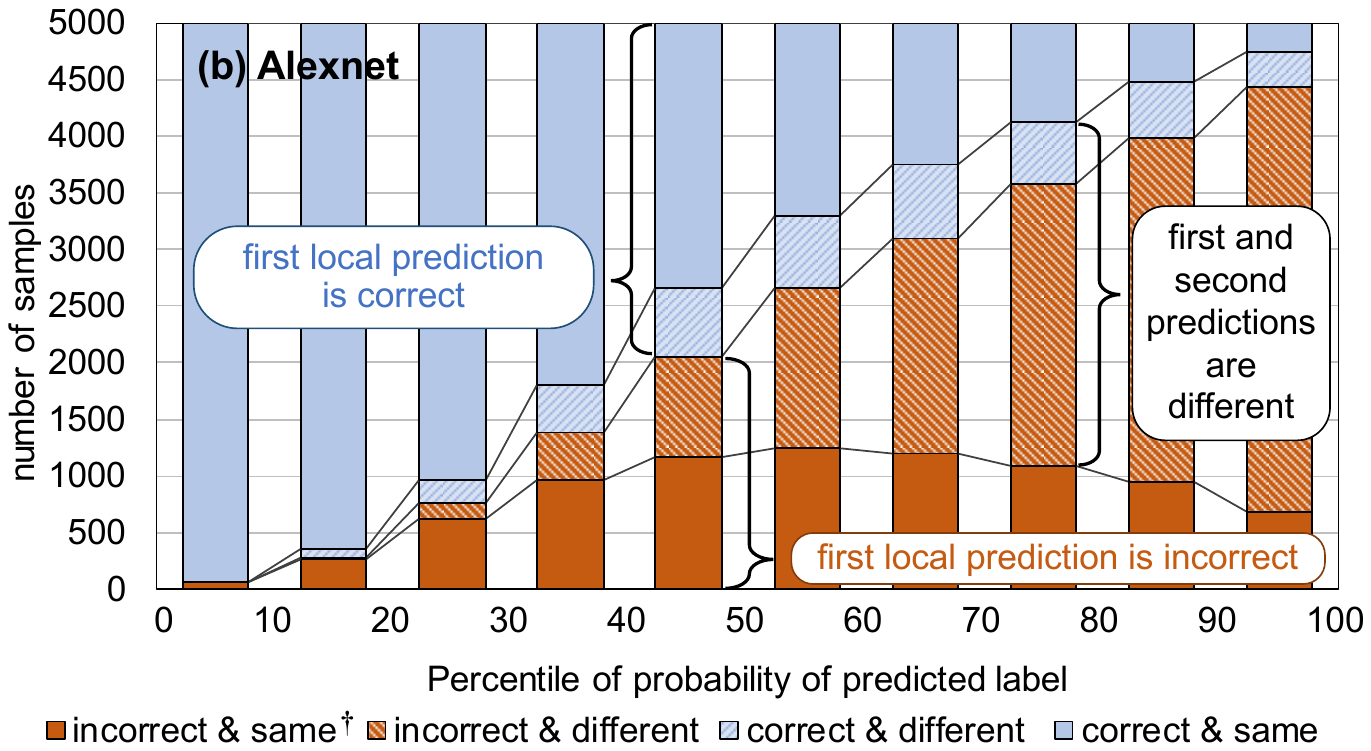}
    \end{minipage}
  \end{tabular}
  \caption{Breakdown of samples into four categories based on 1) whether first local prediction is correct or incorrect and 2) whether first and second predictions are same or different. X-axis shows percentile of probability of first local predictions from high (left) to low (right). Two networks tend to make same (mis)predictions for samples predicted with high probability (left), and, hence, ensembling does not work well for them.
  }

\end{center}
\end{figure*}

In this paper, we first describe our insights on the relationship between the probability of prediction and the effect of ensembling with current deep neural networks; ensembling does not help mispredictions for inputs predicted with a high probability, i.e., the output from the softmax, even when there is a non-negligible number of mispredicted inputs. To exploit this finding to speed up ensembling, we developed adaptive ensemble prediction that maintains the benefits of ensembling with much smaller additional costs. During the ensembling process, we calculate the confidence level of the probability obtained from local predictions for each input. If an input reaches a high enough confidence level, we stop ensembling and making more local predictions for this input to avoid wasting computation power. We evaluated our adaptive ensembling by using four image classification datasets: ILSVRC 2012, CIFAR-10, CIFAR-100, and SVHN. Our results showed that adaptive ensemble prediction reduces the computation cost significantly while achieving accuracy similar to that of static ensemble prediction with a fixed number of local predictions. We also showed that our statistically rigorous confidence-level-based early-exit condition yields a better accuracy with the same cost (or lower cost for the same accuracy) in addition to reducing the burden of task-dependent threshold tuning better compared with a naive early-exit condition based on a pre-defined threshold in the probability.

\section{Ensembling and Probability of Prediction}

This section describes the observations that have motivated us to develop our proposed technique: how ensemble prediction improves the accuracy of predictions with different probabilities.

\subsection{Observations}

To show the relationship between the probability of prediction and the effect of ensembling, we evaluated the prediction accuracy for the ILSVRC 2012 dataset with and without the ensembling of two predictions made by two independently trained networks. Figure 1(a) shows the results of this experiment with GoogLeNet; the two networks follow the design of GoogLeNet and use exactly the same configurations (hence, the differences come only from the random number generator). In the experiment, we 1) evaluated the 50,000 images from the validation set of the ILSVRC 2012 dataset by using the first network without ensembling, 2) sorted the images in terms of the probability of prediction, and 3) evaluated the images with the second network and assessed the accuracy after ensembling two local predictions by using the arithmetic mean. The x-axis of Figure 1(a) shows the percentile of the probability from high to low, i.e., going left (right), and as can be seen, the first local predictions became more (less) probable. The gray dashed line shows the average probability for each percentile class. Overall, ensembling improved accuracy well, although we only averaged two predictions. Interestingly, we can observe that the improvements only appear on the right side of the figure. There were almost no improvements made by ensembling two predictions on the left side, i.e., input images with highly probable local predictions, even when there was a non-negligible number of mispredicted inputs. For example, in the 50- to 60-percentile range with GoogLeNet, the top-1 error rate was 29.6\% and was not improved by averaging two predictions from different networks. 

For more insight into these characteristics, Figure 2(a) shows the breakdown of 5,000 samples in each 10-percentile range into 4 categories based on 1) whether the first prediction was correct or not and 2) whether the two networks made the same prediction or different predictions. When a prediction with a high probability was made first, we can observe that another local prediction tended to be the same regardless of its correctness. In the highest 10-percentile range, for instance, the two independently trained networks made the same misprediction for all 43 mispredicted samples. The two networks made different predictions only for 2 out of the 5,000 samples even when we included the correct predictions. In the 10- to 20-percentile range, the two networks generated different predictions only for 3 out of 139 mispredicted samples. Ensembling does not work well when local predictions tend to make the same mispredictions. 

%However, ensembling does not help mispredictions even with moderate probability; e.g. the average probability of prediction is 79.4\% in the 50- to 60-percentile range. These results show
%it is difficult to change the final output by averaging other local predictions since we select the label with the highest average probability as the final prediction. 

To determine whether or not this characteristic of ensembling is unique to the GoogLeNet architecture, we conducted the same experiment using Alexnet as another network architecture and show the results in Figure 1(b) and 2(b). Although the prediction error rate was higher for Alexnet than for GoogLeNet, we observed similar characteristics of improvements made by ensembling. 

When we combined Alexnet (for the first prediction) and GoogLeNet (the second prediction), ensembling the local prediction from GoogLeNet, which yielded much higher accuracy than the first prediction by Alexnet, did not produce a significant gain in the 0- to 20-percentile range. We discuss this deeper in appendix. Also, we show the results with ResNet50 \cite{He2016} in appendix. Even with ResNet, which has higher accuracy than GoogLeNet, the improvements made by ensembling were only observed on the right side of the figure, i.e., for images with low probabilities. 

These characteristics of the improvements made by ensembling are not unique to an ILSVRC dataset; we have observed similar trends in other datasets.

\subsection{Why This Happens}

To understand why ensembling does not work for inputs predicted with high probabilities, we investigated a simplified classification task, which we detail in appendix, and found a common type of misclassification that ensembling did not help.

Trained neural networks (or any classifiers in general) often make a prediction for a sample near the decision boundary of class A and class B with a low probability; the classifier assigns similar probabilities for both classes, and a class with a higher probability (but with a small margin) is selected as the result. For such samples, ensembling works efficiently by reducing the effects of random perturbations.

While ensembling works near a decision boundary that is properly learned by a classifier, some decision boundaries can be totally missed by a trained classifier due to insufficient expressiveness in the model that is used or a lack of appropriate training data. Figure 3 shows an example of true decision boundaries in 2-D feature space and classification results with a classifier that is not capable of capturing all of these boundaries. In this case, the small region of class A in the top-left was totally missed in the classification results obtained with a trained network, i.e., the samples in this region were mispredicted with high probabilities. Typically, such mispredictions cannot be avoided by ensembling predictions from another classifier trained with different random numbers since these mispredictions are caused by the poor expressiveness of the model rather than the perturbations that come from random numbers. %We show this problem in more detail using an actual neural network for a simple classification task and in the supplemental material.

These results motivated us to make our adaptive ensemble prediction for reducing the additional cost of ensembling while keeping the benefit of improved accuracy. Once we obtain a high enough prediction probability for an input image, further local prediction and ensembling will waste computation power without improving accuracy. The challenge is how to identify the condition under which ensembling is terminated early. As described later, we determine that an early-exit condition based on the confidence level of probability works well for all tested datasets. 

% Also, we show the results when mixing GoogLeNet and Alexnet in the appendix.
\begin{figure}
  \begin{center}
    \includegraphics[width=8cm]{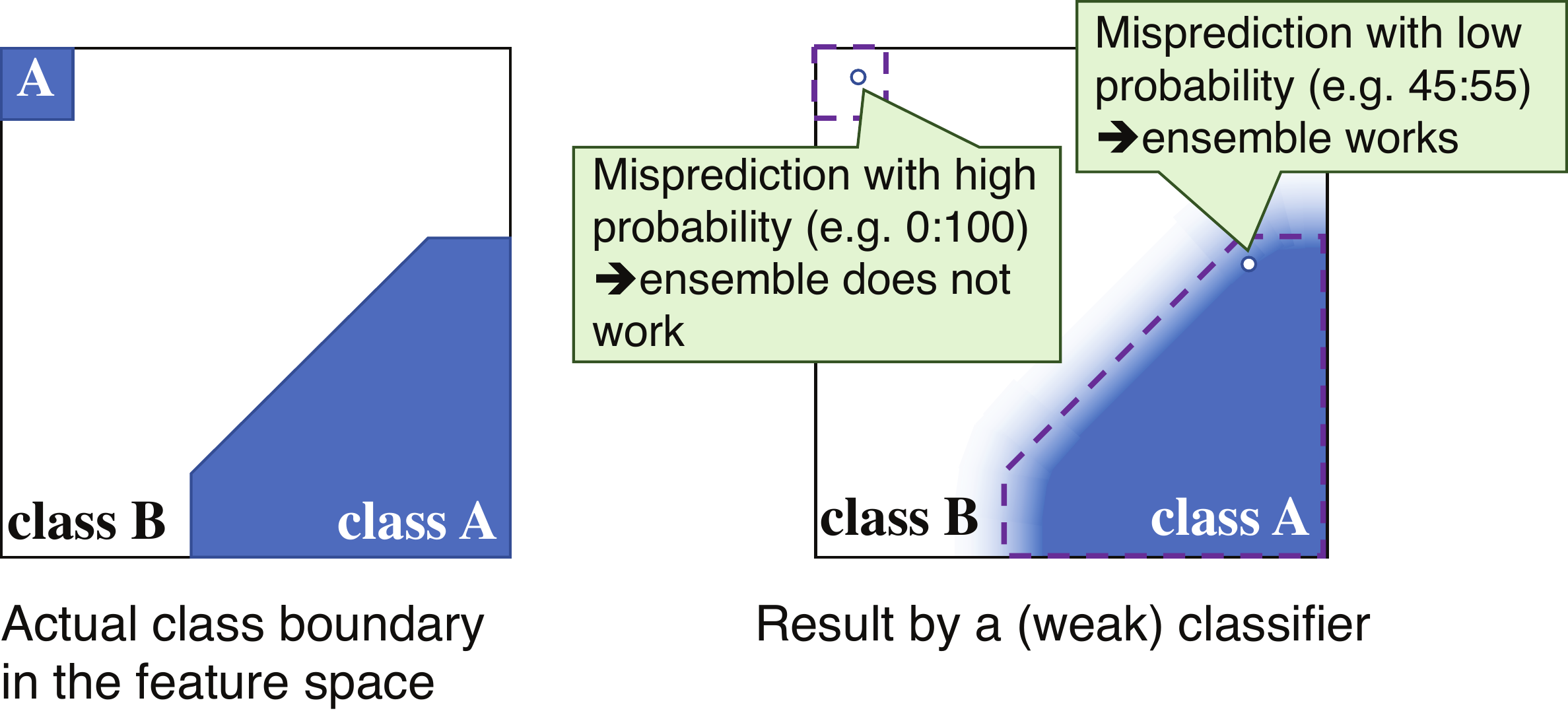}
    \caption{Schematic examples of mispredictions with high and low probabilities. Classifier may fail to learn some decision boundaries, and this leads to mispredictions with high probability; ensembling cannot help them.}
  \end{center}
  \label{boundaries}
\end{figure}

\section{Related Work}

Various prediction methods that ensemble the outputs from many classifiers (e.g., neural networks) have been widely studied to achieve higher accuracy in machine learning tasks. Boosting (Freund and Schapire 1996) and bagging (\cite{Breiman96}) are famous examples of ensemble methods. Boosting and bagging produce enough variances in classifiers included in an ensemble by changing the training set for each classifier. Another technique for improving binary classification using multiple classifiers is soft-cascade (e.g., \cite{Bourdev05}, \cite{Zhang08}). With soft-cascade, multiple weak sub-classifiers are trained to reject a part of negative inputs. Hence, when these classifiers are combined to make one strong classifier, many easy-to-reject inputs are rejected in the early stages without consuming a huge amount of computation time. % Although our technique addresses multi-class classification tasks unlike the original Soft-cascade, our basic insight may potentially useful for extending Soft-cascade (e.g. to use in multi-class classification tasks). Also there are some differences between ours and Soft-cascade; ours is an inference-time technique and does not affect the training phase and ours can be used even with only one network to make efficient ensemble.
Compared with boosting, bagging, or soft-cascade, ours is an inference-time technique and does not affect the training phase.

In recent studies on image classification with deep neural networks, random numbers (e.g., for initialization or for ordering input images) used in the training phase can give sufficient variances in networks even with the same training set for all classifiers (networks). Hence, we use networks trained by using the same training set and network architecture in this study, and we assume that the capabilities of local classifiers are not that different. If a classifier is way too much weaker than the later classifiers as in soft-cascade, the ensembling goes differently compared with our observations as discussed in Section 2; mispredicted inputs in a weak classifier may be predicted correctly by later powerful classifiers even if they are predicted with high probabilities in the local prediction made by the weak classifier. For example, a later classifier that is capable of capturing the top-left region of class A in Figure 3 may predict the sample in the top-left correctly. % If the entire classifier consists of local classifiers with huge performance differences, we need to pay special attention on applying our adaptive ensembling. 

Another series of studies on accelerating classification tasks with two or few classes is based on the dynamic pruning of majority voting (e.g., \cite{Lobato09}, \cite{Soto16}). Like our technique, dynamic pruning uses a certain confidence level to prune  ensembling with a sequential voting process to avoid wasting computation time. We show that the confidence-level-based approach is quite effective at accelerating ensembling by averaging local predictions in many-class classification tasks with deep neural networks when we use the output of the softmax as the probability of the local predictions. COMET by \cite{Basilico2011} stops ensembling for random forest classifiers on the basis of the confidence interval. We also use the confidence interval but in a different way; COMET stops ensembling for binary classification tasks when an unobserved proportion of positive votes falls on the same side of 0.5 as the current observed mean with a certain confidence level. We use the confidence interval to confirm that a predicted label has a higher probability than other labels with confidence. We cannot naively follow GLEE since we do not target binary classification tasks. Also, COMET cannot take our approach because the random forest does not provide probability information for each local prediction unlike neural network classifiers. \cite{Wang2018} decided the order of local predictions and also the thresholds for early stopping by solving a combinatorial optimization problem for binary classification tasks. In our study, we fixed the order of local predictions since our local predictions are based on the same network architecture and ordering is not that important. The basic idea of using the confidence level for the early-exit condition does not depend on this specific order of local predictions. 

Some existing classifiers with a deep neural network (e.g., \cite{Bolukbasi2017}, \cite{Teerapittayanon2017}, \cite{Huang2018}) take an early exit approach in ensembling similar to ours or take an early exit from one neural network. In our study, we study how the early-exit condition affects the execution time and the accuracy in detail and show that our confidence-level-based condition works better than naive threshold-based conditions.

The higher execution cost of ensembling is a known problem, so we are not the first to attack it. For example, \cite{Hinton15} also tackled the high execution cost of ensembling. Unlike us, they trained a new smaller network by distilling the knowledge from an ensemble of networks by following \cite{Bucilua06}. 

To improve the performance of the inference of deep neural networks by making small and efficient executables suitable for handheld devices from trained networks, graph compilers and optimizers, such as NNVM/TVM (\cite{Chen18}) and Glow (\cite{Rotem18}), were recently developed.

In our technique, we use the probability of predictions to control ensembling during inference. Typically, the probability of predictions generated by the softmax is used during the training of a network; the cross entropy of the probabilities is often used as the objective function of optimization. However, using probability for purposes other than the target of optimization is not unique to us. For example, \cite{Hinton15} used probabilities from the softmax while distilling knowledge from an ensemble of multiple models to create a smaller network for deployment. As far as we know, ours is the first study focusing on the relationship between the probability of prediction and the effect of ensembling with deep neural networks. 

\cite{Opitz99} showed an important observation related to ours. They showed that a large part of the gain of ensembling came from the ensembling of the first few local predictions. Our observation discussed in the previous sections enhances Opitz's observation from a different perspective: most of the gain of ensembling comes from inputs with low probabilities in prediction.

\section{Adaptive Ensemble Prediction}

\subsection{Basic Idea}

This section details our proposed adaptive ensemble prediction method. As shown in Figure 1, ensembling typically does not improve the accuracy of predictions if a local prediction is highly probable. Hence, we terminate ensembling without processing all $N$ local predictions on the basis of the probabilities of the predictions. We execute the following steps.

\begin{enumerate}
\item start from $i$ = 1
\item obtain the $i$-th local prediction, i.e., the probability for each class label. We denote the probability for label $L$ of the $i$-th local prediction $p_{L,i}$
\item calculate the average probabilities for each class label 
\begin{equation}
\left<p_L\right>_i = \frac{\sum_{j=1}^{i}p_{L,j}}{i}
\end{equation}

\item if $i < N$, and the early-exit condition is not satisfied, increment $i$, and repeat from step 2
\item output the class label that has the highest average probability $\argmax_{L}\left(\left<p_L\right>_i\right)$ as the final prediction.
\end{enumerate}

\subsection{Confidence-level-based Early Exit}% Condition
For the early-exit condition in step 4, we propose a condition based on confidence level. 

We can use a naive condition on the basis of a pre-determined static threshold $T$ to terminate the ensembling, i.e., we just compare the highest average probability $\max_{L}\left(\left<p_L\right>_i\right)$ against the threshold $T$. If the average probability exceeds the threshold, we do not execute further local predictions for ensembling. As we empirically show later, the best threshold $T$ heavily depends on the task. To avoid this difficult tuning of the threshold $T$, we propose a dynamic and more statistically rigorous condition in this paper.

%\begin{equation}
%\max_{L}\left(\left<p_L\right>_i\right) > T, 
%\end{equation}

Instead of a pre-defined threshold, we can use confidence intervals (CIs) as an early-exit condition. We first find the label that has the highest average probability ({\it predicted label}). Then, we calculate the CI of the probabilities using $i$ local predictions. If the calculated CI of the predicted label does not overlap with the CIs for other labels, i.e., the predicted label is the best prediction with a certain confidence level, we terminate the ensembling and output the predicted label as the final prediction.

We calculate the confidence interval for the probability of label $L$ using $i$ local predictions as follows.
\begin{equation}
\left<p_L\right>_i \pm z \frac{1}{\sqrt{i}}\sqrt{\frac{\sum_{j=1}^{i}\left(p_{L,j}-\left<p_L\right>_i\right)^2}{i-1}}
\end{equation}
Here, $z$ is defined such that a random variable $Z$ that follows the Student's-t distribution with $i-1$ degrees of freedom satisfies the condition $Pr\left[Z \leq z \right] = 1 - \alpha / 2$. $\alpha$ is the significance level, and $(1 - \alpha)$ is the confidence level. We can read the value $z$ from a precomputed table at runtime. To compute the confidence interval with a small number of samples (i.e., local predictions), it is known that the Student's-t distribution is more suitable than the normal distribution. When the number of local predictions increases, the Student's-t distribution approximates the normal distribution.

\begin{figure*}[t]
  \centering
  \includegraphics[clip,width=13cm]{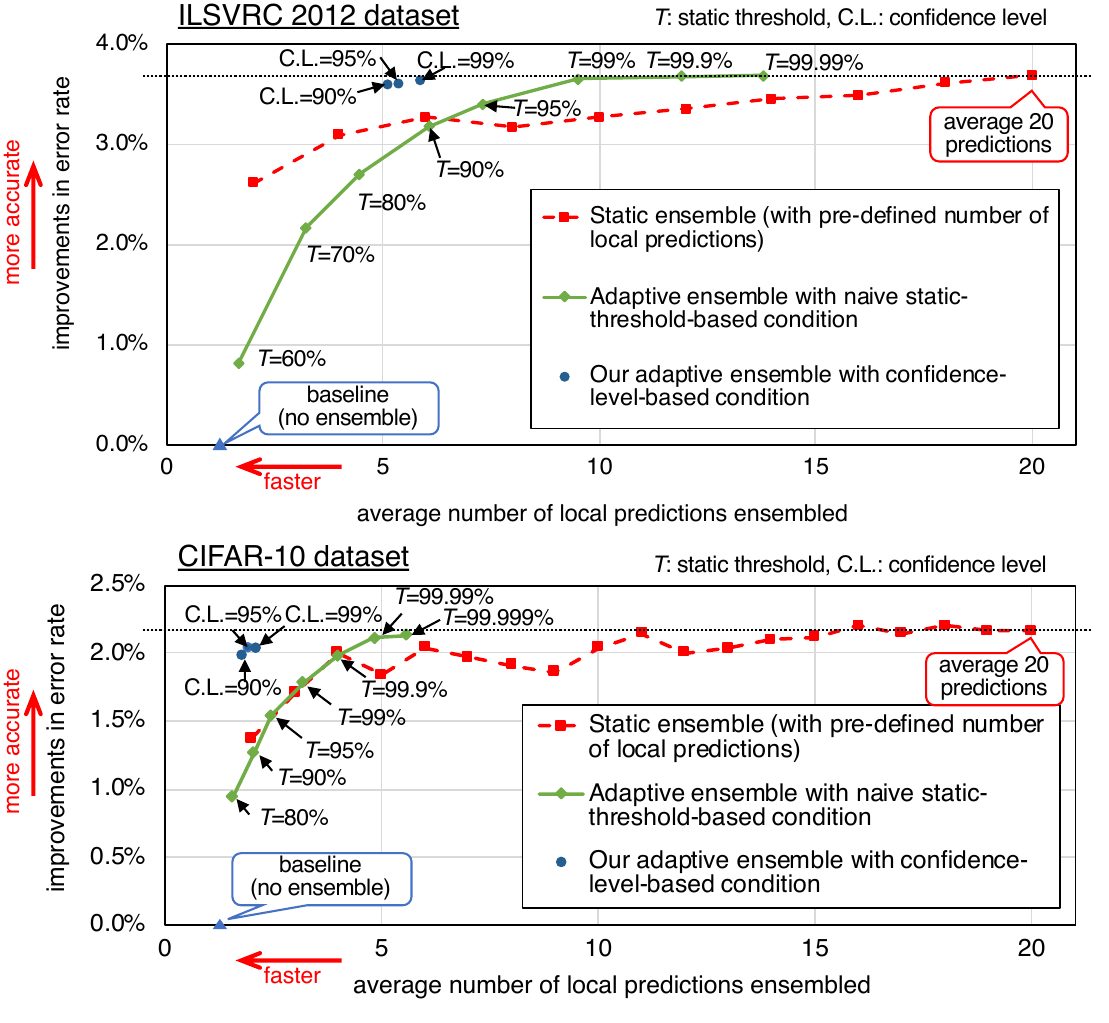}
  \caption{Prediction accuracy and computation cost with static ensemble and our adaptive ensemble using different early-exit conditions. % Static threshold $T$ can be used to control tradeoff between accuracy and computation cost. Static ensemble with all 20 predictions and no ensemble (0.0\% in figure) are two tradeoff extremes. 
  Confidence-level-based condition achieved better accuracy than static-threshold-based conditions with same computation cost, especially for CIFAR-10. Tuning of confidence level (CL) is less sensitive than that of static threshold.}
  \label{tradeoff}
\end{figure*}

We can do pair-wise comparisons between the predicted label and all other labels. However, computing CIs for all labels is costly, especially when there are many labels. To avoid the excess costs of computing CIs, we compare the probability of the predicted label against the total of the probabilities of other labels. Since the total of the probabilities of all labels (including the predicted label) is 1.0 by definition, the total of the probabilities for the labels other than the predicted label is $1 - \left<p_L\right>_i$, and the CI is the same size as that of the probability of the predicted label. Hence, our early-exit condition is as follows.
\begin{equation}
\left<p_L\right>_i - \left(1 - \left<p_L\right>_i\right) > 2z \frac{1}{\sqrt{i}}\sqrt{\frac{\sum_{j=1}^{i}\left(p_{L,j}-\left<p_L\right>_i\right)^2}{i-1}}
\end{equation}
We avoid computing CI if $\left<p_L\right>_i < 0.5$ to avoid wasteful computation because the early-exit condition of equation 2 cannot be met in such cases. Since the CI cannot be calculated with only one local prediction as is obvious from equation (3) to avoid zero divisions, we can use a hybrid of the two early-exit conditions. We use the static-threshold-based condition only for the first local prediction with a quite conservative threshold (99.99\% in the current implementation) to terminate ensembling only for trivial inputs as early as possible, and after the second local prediction is calculated, the confidence-level-based condition of equation (3) is used. We also performed evaluation by doing pair-wise comparisons of CIs with all other labels or with the label having the second-highest probability. However, the differences in the results obtained by doing pair-wise comparisons were mostly negligible. There can be many other criteria for the early-exit conditions, but our approach with the confidence level is a reasonable choice for balancing multiple objectives, including accuracy, computation cost, and ease of tuning.

\section{Experiments}

\begin{table*}[t]
  \centering
  \caption{Prediction accuracy with and without adaptive ensemble}
  \includegraphics[clip,width=13cm,trim=0 7cm 0 0]{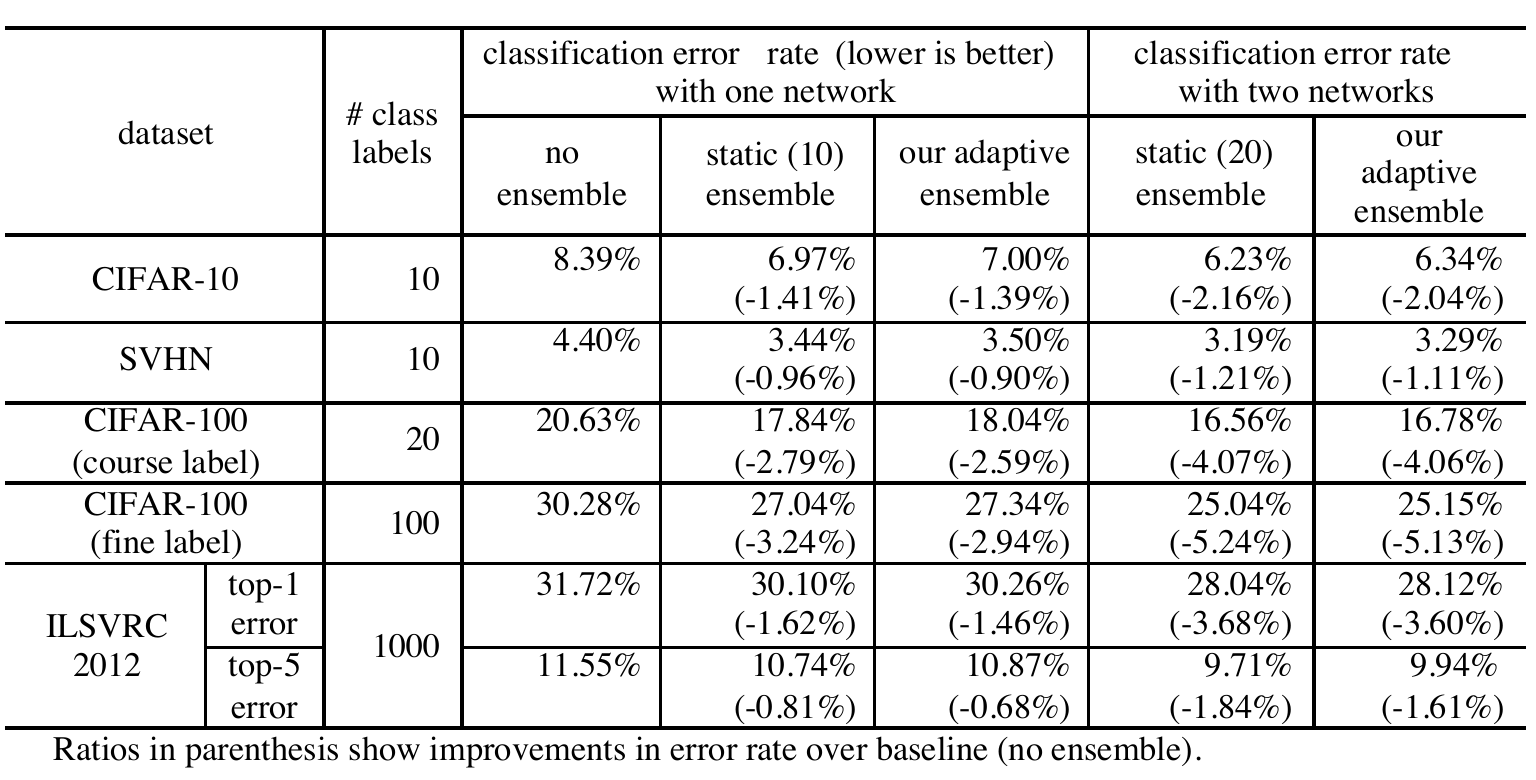}

  \centering
  \caption{Number of local predictions ensembled with and without adaptive ensemble}
  \includegraphics[clip,width=14.0cm,trim=0 13.5cm 0 0]{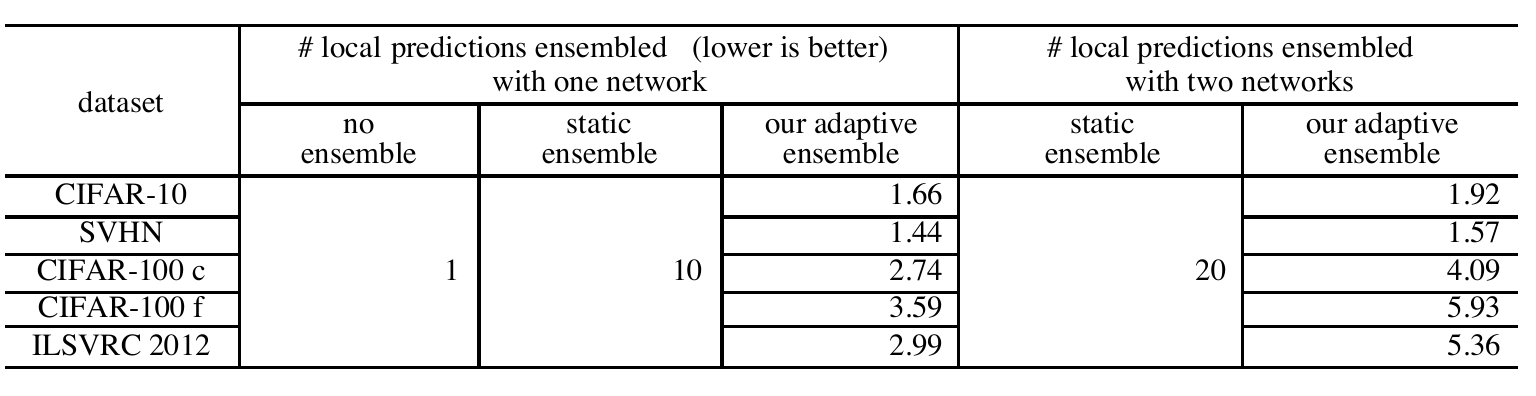}

  % \centering
  % \caption{Execution time with and without adaptive ensemble}
  % \includegraphics[clip,width=12.0cm,trim=0 10cm 0 0]{table2.pdf}
\end{table*}

\subsection{Implementation}

In this section, we investigate the effects of adaptive ensemble prediction on the prediction accuracy and execution cost with various image classification tasks: ILSVRC 2012, Street View House Numbers (SVHN), CIFAR-10, and CIFAR-100 (with fine and coarse labels) datasets.

For the ILSVRC 2012 dataset, we used GoogLeNet as the network architecture and trained the network by using stochastic gradient descent with momentum as the optimization method. For other datasets, we used a network that has six convolutional layers with batch normalization (\cite{Ioffe15}) followed by two fully connected layers with dropout. We used the same network architecture except for the number of neurons in the output layer. We trained the network by using Adam (\cite{Kingma15}) as the optimizer. For each task, we trained two networks independently. During the training, we used data augmentations by extracting a patch from a random position of an input image and using random horizontal flipping. Since adaptive ensemble prediction is an inference-time technique, network training is not affected. 

We averaged up to 20 local predictions by using ensembling. We created 10 patches from each input image by extracting from the center and four corners of images with and without horizontal flipping by following Alexnet. For each patch, we made two local predictions using two networks. The patch size was 224 x 224 for the ILSVRC 2012 dataset and 28 x 28 for the other datasets. We made local predictions in the following order: (center, no flip, network 1), (center, no flip, network 2), (center, flipped, network 1), (center, flipped, network 2), (top-left, no flip, network 1), ..., (bottom-right, flipped, network 2). Since averaging local predictions from different networks typically yields better accuracy, we used this order for both our adaptive ensembling and fixed-number static ensembling. As far as we tested, the order of local predictions slightly affected the error rates, but it did not change the overall comparisons shown in the evaluations.

\subsection{Results}

To study the effects of our adaptive ensembling on the computation cost and accuracy, we show the relationship between them for the ILSVRC 2012 and CIFAR-10 datasets in Figure \ref{tradeoff}. We used two networks in this experiment, i.e., up to 20 predictions were ensembled. In the figure, the x-axis is the number of ensembled predictions, so smaller means faster. The y-axis is the improvements in classification error rate over the baseline (no ensemble), so higher means better. We evaluated the static ensembling (averaging the fixed number of predictions) by changing the number of predictions and our adaptive ensembling. For the adaptive ensembling, we also performed evaluation with two early-exit conditions: with naive static threshold and with confidence interval. We tested the static-threshold-based condition by changing the threshold $T$ and drew lines in the figure. Similarly, we evaluated the confidence-level-based condition with three confidence levels frequently used in statistical testing: 90\%, 95\%, and 99\%. 

From the figure, there is an obvious trade-off between the accuracy and the computation cost. The static ensemble with 20 predictions was at one end of the trade-off because it never exited early. The baseline, at which ensembling was not executed, was at the other end, and it always terminated at the first prediction regardless of the probability. Our adaptive ensembling with confidence-level-based condition achieved better accuracy with the same computation cost (or smaller cost for the same accuracy) compared with the static or naive adaptive ensembling with a static threshold. %Especially, the adaptive ensemble largely outperformed the static ensemble for both datasets.
The gain with the confidence-level-based condition over the static-threshold-based was significant for CIFAR-10, whereas it was marginal for ILSVRC 2012. These two datasets showed the largest and smallest gain with the confidence-level-based condition over the static-threshold-based condition; the other datasets showed improvements between those of the two datasets shown in Figure \ref{tradeoff}. 

When comparing the two early-exit conditions in adaptive ensembling, the confidence-level-based condition eliminated the burden of parameter tuning better compared with the naive threshold-based condition in addition to the benefit of the reduced computation cost. Obviously, how to decide the best threshold $T$ is the most important problem for the static-threshold-based condition. The threshold $T$ can be used as a knob to control the trade-off between accuracy and computation cost, but the static threshold tuning is highly dependent on the dataset and task. From Figure \ref{tradeoff}, for example, $T=90\%$ or $T=95\%$ seem to have been a reasonable choice for ILSVRC 2012, but it was a problematic choice for CIFAR-10. For the confidence-level-based condition, the confidence level also controlled the trade-off. However, the differences in the computation cost and the improvements in accuracy due to the choice of the confidence level were much less significant and less sensitive to the current task than the differences due to the static threshold. Hence, task-dependent fine tuning of the confidence level is not as important as the tuning of the static threshold. The easier (or no) tuning of the parameter is an important advantage of the confidence-level-based condition.

%the threshold $T$ can be used as a knob to control the tradeoff between the accuracy and the computation cost as well as the number of predictions on average in the naive ensemble. For both our adaptive ensemble with the static threshold and the naive ensemble, By comparing the two lines, the adaptive ensemble with the static threshold achieved accuracy better than or comparable to the naive ensemble using the same number of predictions on average unless the threshold $T$ was too small. This means that the probability of prediction is an effective criterion to control the number of predictions to ensemble. From Figure 1, it is reasonable that an excessively small threshold $T$, e.g. less than 80\%, decreases the accuracy since it will significantly miss the opportunity that we can gain from ensembling.

Tables 1 and 2 show how adaptive ensemble prediction affected the accuracy of predictions and the execution costs in more detail for five datasets. Here, for our adaptive ensembling, we used the confidence-level-based early-exit condition with a 95\% confidence level for all datasets on the basis of the results of Figure \ref{tradeoff}. We tested two different configurations: with one network (i.e., up to 10 local predictions) and with two networks (up to 20 local predictions). In all datasets, the ensembling improved the accuracy in the trade-off for the increased execution costs as expected. Using two networks doubled the number of local predictions on average (from 10 to 20) and increased both the benefit and drawback. If we were to use further local predictions (e.g., original GoogLeNet averaged up to 1,008 predictions), the benefit and cost would become much more significant. Comparing our adaptive ensembling with static ensembling, our adaptive ensembling similarly improved accuracy while reducing the number of local predictions used in the ensembles; the reductions were up to 6.9 and 12.7 times for the one-network and two-network configurations.
%The reduced numbers of local predictions result in shorter execution time; the speedup was by 2.1x to 2.8x and by 2.3x to 3.5x for the one-network and two-network configurations, respectively. The reductions in the execution time over the static ensemble are smaller than the reduction in the number of averaged predictions because of the additional overhead due to the confidence interval calculation, which is written in Python in the current implementation. Also, mini batches gradually become small as ensembling for parts of inputs terminated. The smaller batch sizes reduce the efficiency of execution on current GPUs.
Since the speed-up with our adaptive technique over static ensembling becomes larger as the number of max predictions to ensemble increases, the benefit of our adaptive technique will become more impressive if we use larger ensemble configurations.

\section{Conclusion}

In this paper, we described our adaptive ensemble prediction with statistically rigorous confidence-level-based early-exit condition to reduce the computation cost of ensembling many predictions. We were motivated to develop this technique by our observation that ensembling does not improve the prediction accuracy if predictions are highly probable. Our experiments using various image classification tasks showed that our adaptive ensembling makes it possible to avoid wasting computing power without significantly sacrificing prediction accuracy by terminating ensembles on the basis of the probabilities of the local predictions. The benefit of our technique will become larger if we use more predictions in an ensemble. Hence, we expect our technique to make ensemble techniques more valuable for real-world systems by reducing the total computation power required while maintaining good accuracy and throughput.

\bibliographystyle{apalike}
\bibliography{aistats2019}

\section{Appendix}

Here, we show additional results and discussion on mispredictions with high probabilities.

\subsection{How mispredictions with high probability happen}
\begin{figure*}[t]
  \begin{center}
    \includegraphics[width=14cm]{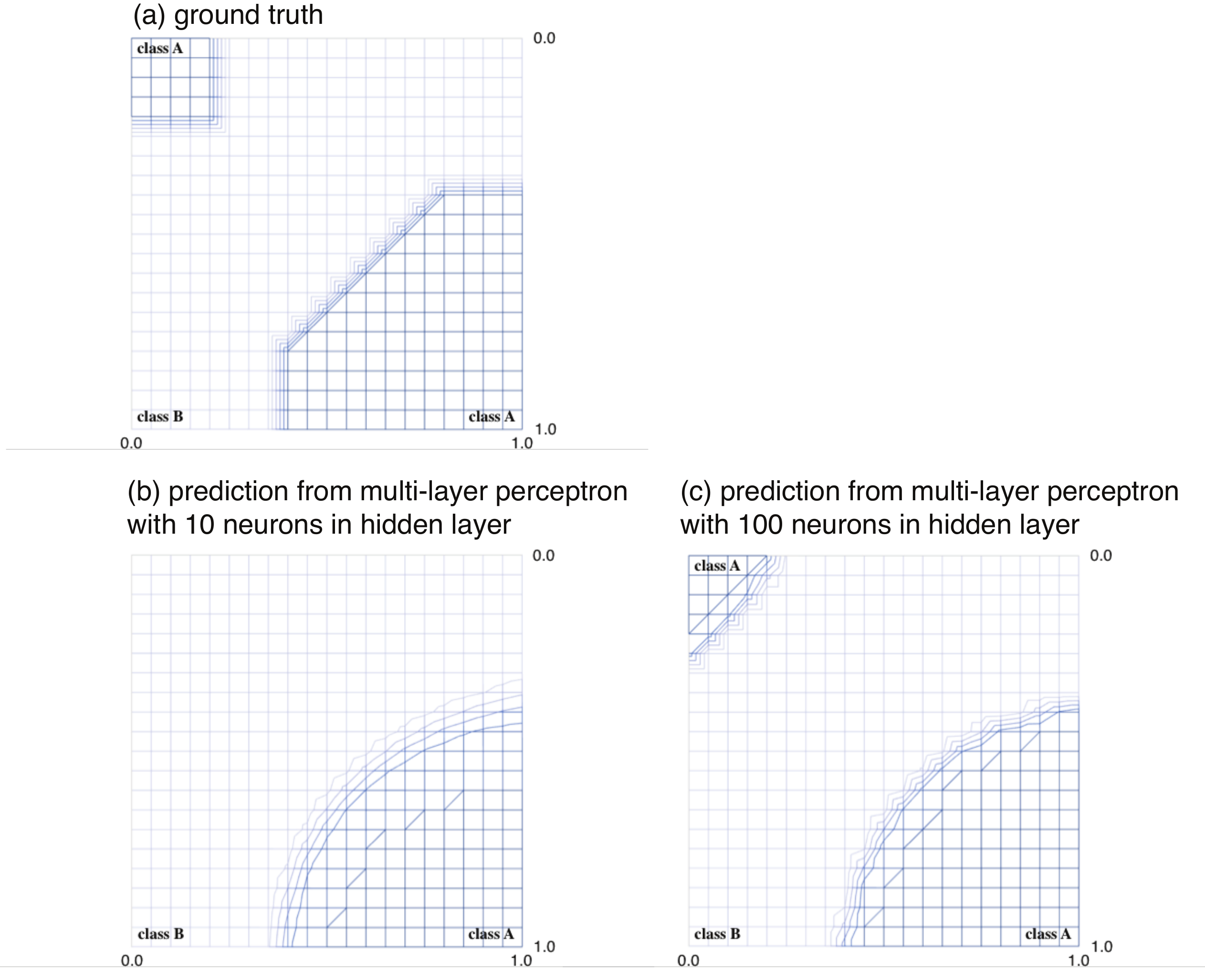}
    \caption{An example of mispredictions with high probabilities due to limited expressiveness of the classifier. A classifier with limited capability (10 hidden neurons) fails to learn the decision boundaries at the top-left region, while a classifier with higher capability (100 hidden neurons) can capture this decision boundary. The weak classifier makes incorrect classifications with almost 100\% probabilities in the top-left region.}
  \end{center}
\end{figure*}

In Section 2.2, we discussed that the mispredictions with high probabilities can be caused by the insufficient expressiveness in the used model as one possible reason.

We show a simple example in Figure 5. We built a three-layer perceptron for a simple binary classification problem which maps a 2-D feature into class A or B as shown in Figure 5(a). We label a sample with feature $(x, y)$, where $0 \leq x < 1$ and $0 \leq y < 1$ as 
\[
  label(x,y) = \left\{ \begin{array}{ll}
    class A & (x \leq 0.2, y \leq 0.2) \\
    class A & (x+y \geq 1.2, x \geq 0.4, y \geq 0.4) \\
    class B & (otherwise)
  \end{array} \right.
\]
 We use the three-layer perceptron with 10 or 100 neurons in the hidden layer as the classifier and Sigmoid function as the activation function. The output layer has only one neuron and the classification results are shown as high or low value. We generated 1,000 random samples as the training data. The training is done by using stochastic gradient descent as the optimizer for 10,000 epochs.

Figure 5(b) depicts the classification results with 10 hidden neurons. In this case, the top-left region of class A is not captured by the classifier at all due to the poor expressiveness of the network even though we have enough training samples in this region. As a result, this (weak) classifier misclassifies the samples in this region for the class B with almost 100\% probability. Although we repeat the training using different random number seeds, this mispredictions cannot be avoided with 10 hidden neurons. Hence, the ensembling multiple local predictions from this classifier cannot help this type of mispredictions in the top-left region. The decision boundary between class A and B at the bottom-right region is not sharp and its shape differs run by run due to the random numbers. The ensembling can statistically reduce the mispredictions near the boundary by reducing the effects of the random numbers.

When we increase the number of hidden neurons from 10 to 100, the top-left region of class A is captured as shown in Figure 5(c). So the expressiveness of the used model matters to avoid the mispredictions with high probabilities.

Another type of the mispredictions with high probabilities can happen if we do not have enough training data in small regions, e.g. the top-left region of class A in the above example. In such case, of course, even highly capable classifiers cannot learn the decision boundary around the small region and mispredict the samples in this region with high probability.

Ensembling multiple local predictions from the multiple local classifiers does not help both types of mispredictions and hence stop ensembling for them is effective to avoid wasting computation power without increasing the overall error rate.

\subsection{Ensembling and Probability of Prediction}

\subsubsection{Results with ResNet50}

In Section 2 of the paper, we showed the relationship between the probability of prediction and the effect of ensembling using GoogLeNet and Alexnet. To show that our observations are still valid with newer network architectures, the result with ResNet50 \cite{He2016} is shown in Figure 6. In addition to the random cropping and flipping data augmentation used for experiments with GoogLeNet and Alexnet, we also employ sample pairing data augmentation technique \cite{Inoue2018} to achieve further improvements in accuracy. Our observation, ensembling does not help mispredictions for inputs predicted with a high probability, is still valid for a newer network architecture and with more advanced data augmentation technique as we can see by comparing the results for ResNet (Figure 6) and GoogLeNet and Alexnet (Figure 1).

\begin{figure}
  \begin{center}
    \includegraphics[width=8cm]{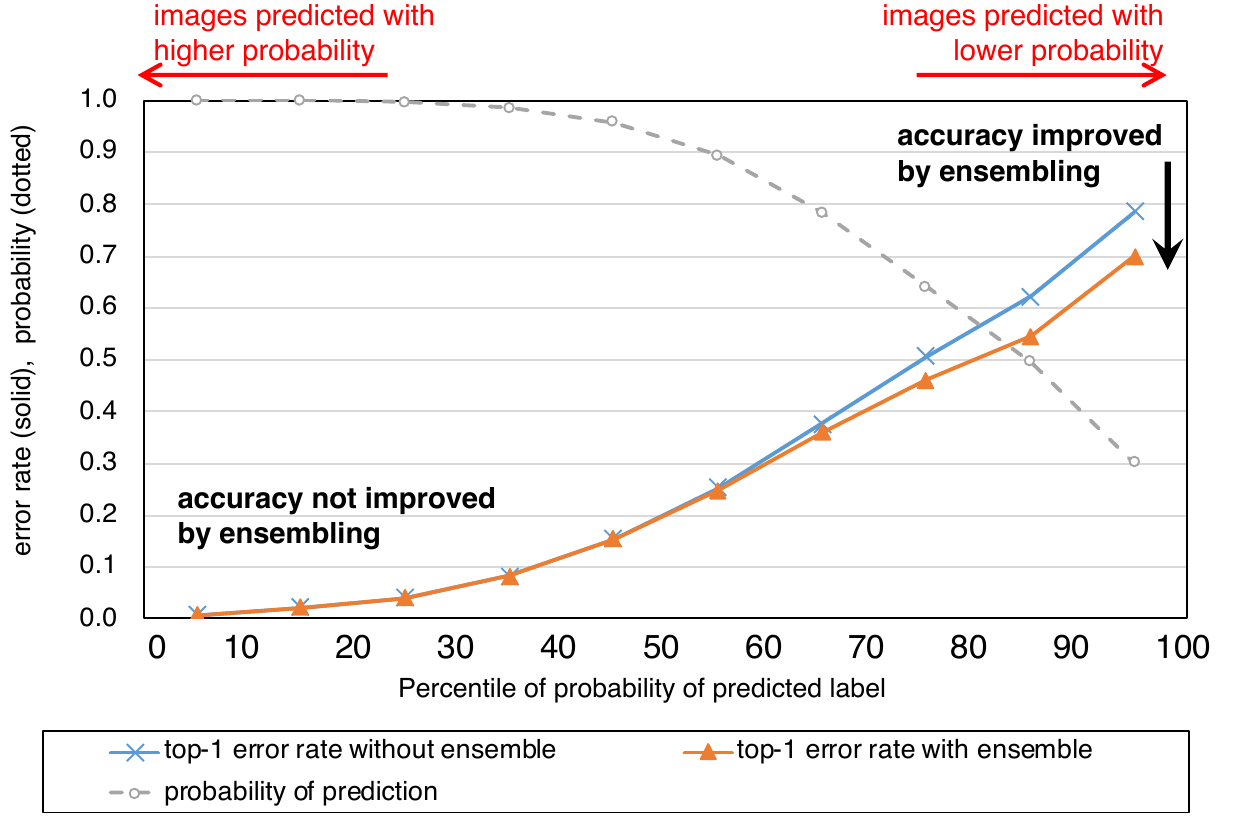}
    \caption{Improvements by ensemble and probabilities of predictions in ILSVRC 2012 validation set using ResNet50. X-axis shows percentile of probability of first local predictions from high (left) to low (right).}
  \end{center}
\end{figure}

\begin{figure*}
  \begin{center}
  \begin{tabular}{c}
  \setlength{\tabcolsep}{10mm}
    \begin{minipage}{0.48\hsize}  
      \includegraphics[width=8cm]{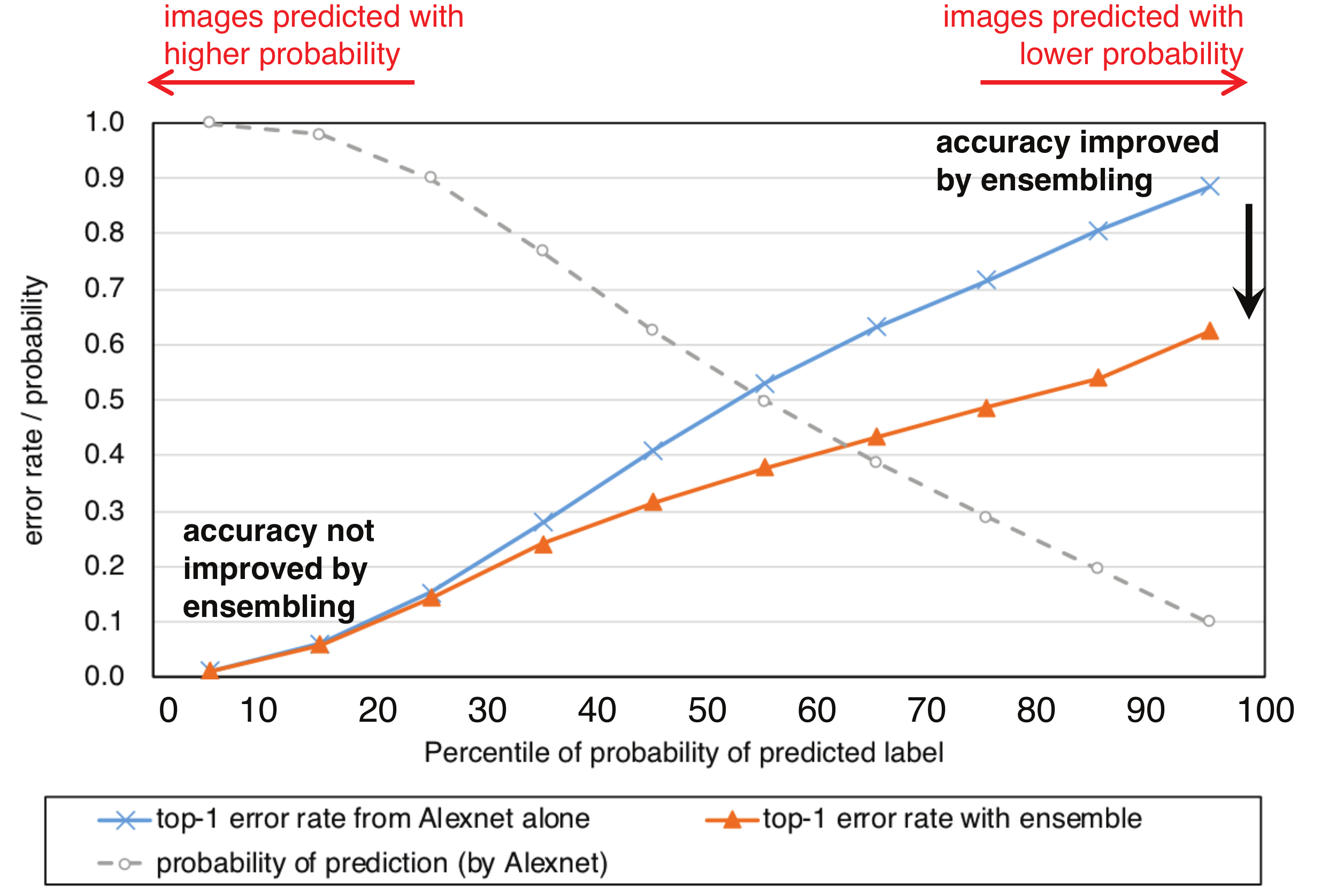}
      \caption{Improvements by ensemble and probabilities of predictions in ILSVRC 2012 validation set using Alexnet and GoogLeNet. X-axis shows percentile of probability of first local predictions from high (left) to low (right). Ensemble reduces error rates for inputs with low probabilities but does not affect inputs with high probabilities.}
    \end{minipage}
    \begin{minipage}{0.1\hsize}
    \end{minipage}
    \begin{minipage}{0.48\hsize}  
      \includegraphics[width=8cm]{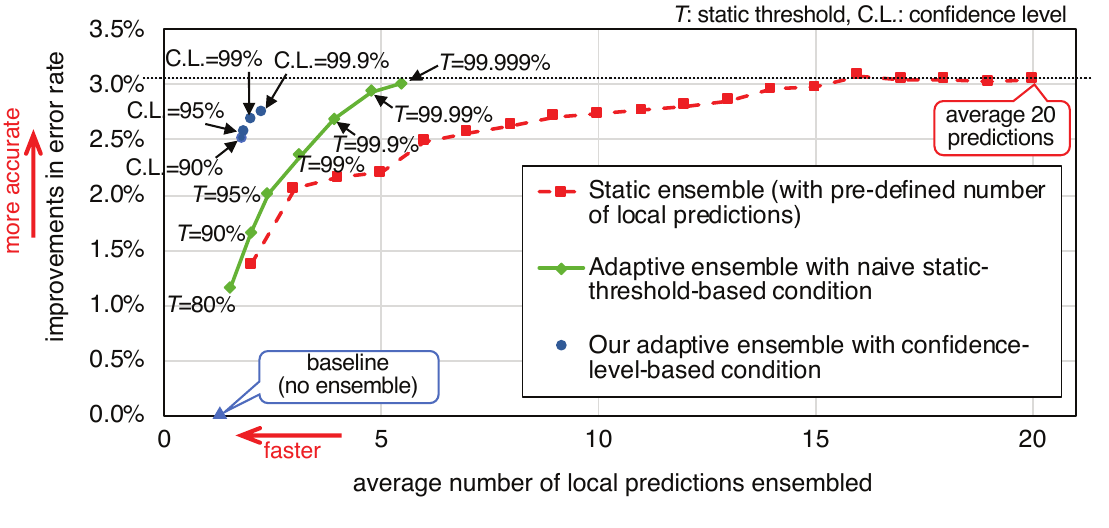}
      \caption{Prediction accuracy and computation cost for CIFAR-10 with static ensemble and our adaptive ensemble using different early-exit conditions. Here we use independently-trained network for each of the local prediction (hence with up to 20 networks in total). }
    \end{minipage}
  \end{tabular}
  \end{center}
\end{figure*}

\subsubsection{Ensemble using different networks}

In Section 2, we showed that ensembling two predictions from two local classifiers of the same network architecture (Alexnet \cite{Krizhevsky12} or GoogLeNet \cite{Szegedy15}) can improve the prediction accuracy only for samples that have low probabilities of prediction. Here, we show the results when we mix the predictions from Alexnet and GoogLeNet. Figure 7 shows the result when we use Alexnet in the first prediction and GoogLeNet in the second. The x-axis shows the percentile of the probability of the prediction by Alexnet from high to low as in figures shown in the main paper. The basic characteristics with two different networks are consistent with the cases using two identical networks discussed in the paper, although the improvements from the ensemble is much more significant since the second local classifier (GoogLeNet) is more powerful than the first one (Alexnet). For the leftmost region, i.e. 0- to 20-percentile samples, the ensemble from the two different networks does not improve the accuracy over the results with only the first local classifier. For the rightmost region, the ensemble improves the error rate significantly.

Here, the ensemble improves the accuracy for much wider regions compared to the cases with two identical networks. For the 20- to 40-percentile range, ensembling two local predictions from Alexnet does not improve the accuracy as shown in Figure 1(b) while ensembling local predictions from Alexnet and GoogLeNet yields improvements as shown in Figure 7. As discussed above using Figure 5(b) and 5(c), a more powerful classifier can avoid some mispredictions with high probabilities. GoogLeNet, which has higher capability than Alexnet, can correctly classify some samples that are misclassified with high probabilities by Alexnet in this range. However, GoogLeNet cannot do better classification in the 0- to 20-percentile range compared to Alexnet. 

\subsection{Prediction Accuracy and Computation Cost with 20 networks }

In Section 4, we showed our evaluations using two networks and ten patches for each of the network. Here, we show the result of evaluation for CIFAR-10 using 20 independently-trained networks for 20 local predictions in Figure 8. The all local predictions use the same input image, which is extracted from the center of the input image without horizontal flipping. 
From the figure, the improvements from ensemble are more significant than the improvements with two networks because local predictions from different models are less correlated each other than the local predictions from the same network. Even in this configuration, our adaptive ensemble with the proposed confidence-level-based condition achieves better accuracy than static-threshold-based conditions with the same number of average evaluations. 

% \begin{figure}[t]
%   \begin{center}
%     \includegraphics[width=8cm]{CIFAR10_20models.pdf}
%     \caption{Prediction accuracy and computation cost for CIFAR10 with static ensemble and our adaptive ensemble using different early-exit conditions. Here we use independently-trained network for each of the local prediction (hence with up to 20 networks in total). }
%   \end{center}
% \end{figure}

\end{document}